\documentclass{article}

    \PassOptionsToPackage{numbers, compress}{natbib}
\usepackage[preprint]{neurips_2020}

\usepackage[utf8]{inputenc} % allow utf-8 input
\usepackage[T1]{fontenc}    % use 8-bit T1 fonts
\usepackage{hyperref}       % hyperlinks
\usepackage{url}            % simple URL typesetting
\usepackage{booktabs}       % professional-quality tables
\usepackage{amsfonts}       % blackboard math symbols
\usepackage{nicefrac}       % compact symbols for 1/2, etc.
\usepackage{microtype}      % microtypography

\usepackage{amsmath}
\usepackage{multirow}
\usepackage{graphicx}

\usepackage{caption}
\usepackage{subcaption}

\title{Momentum Calibration for Text Generation}

\author{%
 Xingxing Zhang$^{1}$~\thanks{Equal contribution.} , Yiran Liu$^{2}$$^*$, Xun Wang$^{1}$, Pengcheng He$^{1}$ \\
 {\bf Yang Yu$^{2}$, Si-Qing Chen$^{1}$, Wayne Xiong$^{1}$, Furu Wei$^{1}$} \\
 $^{1}$Microsoft \\
 $^{2}$Tsinghua University \\
}

\begin{document}

\maketitle

\begin{abstract}
The input and output of most text generation tasks can be transformed to two sequences of tokens and they can be modeled using sequence-to-sequence learning modeling tools such as Transformers. These models are usually trained by maximizing the likelihood the output text sequence and assumes the input sequence and all \emph{gold} preceding tokens are given during training, while during inference the model suffers from the \emph{exposure bias} problem (i.e., it only has access to its previously predicted tokens rather gold tokens during beam search). In this paper, we propose MoCa ({\bf Mo}mentum {\bf Ca}libration) for text generation. MoCa is an online method that dynamically generates slowly evolving (but consistent) samples using a momentum moving average generator with beam search and MoCa learns to align its model scores of these samples with their actual qualities. Experiments on four text generation datasets (i.e., CNN/DailyMail, XSum, SAMSum and Gigaword) show MoCa consistently improves strong pre-trained transformers using vanilla fine-tuning and we achieve the state-of-the-art results on CNN/DailyMail and SAMSum datasets.

\end{abstract}

\section{Introduction}
Text generation is the task of generating a text sequence given certain input and the input can usually be transformed into another text sequence, which is a typical Sequence-to-Sequence (Seq2Seq) learning problem \cite{sutskever2014sequence}.
Large pre-trained Sequence-to-Sequence Transformer models such as T5 \cite{raffel2020exploring}, BART \cite{lewis-etal-2020-bart} and PEGASUS \cite{zhang:2019:icml} have become the default modeling tools for text generation due to its impressive results on a wide range of generation tasks such as summarization \cite{lewis-etal-2020-bart,zhang:2019:icml}, data/keywords-to-text generation \cite{lin2019commongen} as well as machine translation \cite{liu2020multilingual}. These large models are firstly pre-trained on large scale unlabeled datasets and then fine-tuned on labeled datasets for specific tasks.

The dominant approach for fine-tuning is to maximize the likelihood of gold output sequences (MLE; Maximum Likelihood Estimation). By applying the chain rule, it essentially maximizes the probability of each token in the output sequence given all \emph{gold} preceding tokens and the input sequence. The training loss above is at word level. However, during test time, the model needs to predict the whole output sequence from scratch (greedily) using beam search (greedy search can be viewed as beam search with beam size of one). Unlike during training, the model only has access to its own predictions (rather than the \emph{gold} prefix). This discrepancy between training and inference is called \emph{exposure bias} \cite{ranzato:2015:arxiv} (i.e., the model is never exposed to its own prediction errors \cite{wiseman-rush-2016-sequence}). Another problem due to the training and inference discrepancy is the \emph{loss-evaluation mismatch} \cite{wiseman-rush-2016-sequence}, where during training the token-level MLE loss is employed, while during inference sequence level metrics (e.g., BLEU \cite{papineni-etal-2002-bleu}) are usually used.

To address the issues above, samples from the model distribution are needed during training time, so that potential errors in these samples can be ``corrected'' as training progress. \cite{ranzato:2015:arxiv,edunov2017classical} propose to leverage reinforcement learning to guide model training by viewing BLEU \cite{papineni-etal-2002-bleu} or ROUGE \cite{lin-2004-rouge} scores of generated samples as rewards. In scheduled sampling \cite{bengio2015scheduled}, some tokens in the gold target sequence are replaced with model predictions in the later stage of MLE training. However, we argue all these methods above still cannot fix the discrepancy between training and inference. Because samples from a model in \cite{ranzato:2015:arxiv,edunov2017classical,bengio2015scheduled} are treated in isolation, while in beam search we compare multiple hypotheses in each step (and select the top-$K$). In other words, \emph{relative} qualities of different samples are more important than their \emph{absolute} qualities during beam search. Therefore, the probabilities of our generated samples should be aligned with their actual quality (i.e., samples of higher quality should be assigned to higher probabilities by our model). Secondly, our method should be \emph{online} to ensure that samples from our model can represent its model distribution in real time during training. At the same time, these samples should be generated from a similar model so that they are consistent in styles, which may help make the learning process easier. To this end, we use a momentum model to generate samples.

With these design principles, we propose MoCa ({\bf Mo}mentum {\bf Ca}libration)  for Text Generation, which calibrates a model trained with the MLE loss by aligning probabilities of its samples with their quality. In MoCa, we have a \emph{generator} model and an \emph{online} model. The \emph{generator}, which is a momentum moving average of the \emph{online} model, generates slowly evolving samples. These samples are then evaluated with an evaluation model and their quality are estimated. We finally align the evaluation scores with the \emph{online} model scores using a ranking loss. To further minimizing the discrepancy between training and inference, we propose new \emph{online} model scoring functions tailored for beam search. Experiments on four text generation datasets (i.e., CNN/DailyMail, XSum, SAMSum and Gigaword) show MoCa consistently improves strong pre-trained transformers using vanilla fine-tuning and we obtain the state-of-the-art results on CNN/DailyMail and SAMSum.

\section{Related Work}
A Seq2Seq model is typically trained by the word level MLE loss, while during test time, the model predicts the next token based on its own previous predictions using beam search. To address this, the method of showing the model its own predictions during training is first explored in the context of structure prediction \cite{daume2009search}. In sequence-to-sequence learning, scheduled sampling \cite{bengio2015scheduled} propose to replace some of gold tokens in target sequences with its own model predictions during later stage of training. \cite{ranzato:2015:arxiv,edunov2017classical,bahdanau2016actor} generate whole candidate target sequences during training and the generated sequences are viewed as action sequences in reinforcement learning (RL) using BLEU or ROUGE as rewards. Their models are optimized using the REINFORCE algorithm \cite{williams1992simple}, since BLEU or ROUGE scores are non-differentiable. \cite{wiseman-rush-2016-sequence} introduce a method to optimize the beam search process during training by encouraging the gold prefix to appear in the beam with a margin-base loss. Candidate samples in these method above are viewed in isolation and high model scores are assigned to samples which are more similar to their gold samples. While in our method, we compare multiple samples given the same input and encourage our model to assign higher scores to samples with better quality. Besides, our method is differentiable and the optimization challenge in  REINFORCE can be avoided.

% ranking based methods?
Contrastive learning has been applied to text generation \cite{pan-etal-2021-contrastive,cao-wang-2021-cliff,xu2022sequence,cho-etal-2021-contrastive}. They also generate multiple candidate samples for each input, but these samples are used as \emph{hard} negative examples and the positive example is the gold output sequence. The contrastive objective aims to assign higher model scores to positive examples while lower scores to negative examples. Unlike our method, the relative model scores of negative examples are not modeled, which is important for beam search.

Our method is also related to two-stage re-ranking based methods for text generation \cite{shen-etal-2004-discriminative,och-etal-2004-smorgasbord,wan2015multi,mizumoto-matsumoto-2016-discriminative,lee-etal-2021-discriminative,liu-liu-2021-simcls}, since our model attempts to fix the mismatch between model scores of candidate samples and their quality using a ranking objective. \cite{liu-liu-2021-simcls} propose to re-ranking candidates from a neural text generation model based on BART \cite{lewis-etal-2020-bart} (or PEGASUS \cite{zhang:2019:icml}) using a RoBERTa \cite{liu2019roberta} based re-ranker. Different from the methods above, our re-ranker and our text generation model share model parameters.  \cite{liu-etal-2022-brio} and \cite{zhao2022calibrating} use a similar ranking objective as our method and parameters of their generation model and re-ranker are also shared. However, both \cite{liu-etal-2022-brio} and \cite{zhao2022calibrating} are offline methods and their candidate samples are fixed during training and these samples may become \emph{out-of-date} as training progress. With a momentum updated generator, our method can generate \emph{slow evolving} candidate samples that can represent the modeling capability of the current model during the whole training process. Besides, we also find the log probabilities used in model scoring is sub-optimal and we propose new scoring functions tailored for beam search decoding.

\section{Model}
\label{sec:model}
% I will start from this Section.

\begin{figure}
    \centering
    \includegraphics[width=0.85\textwidth]{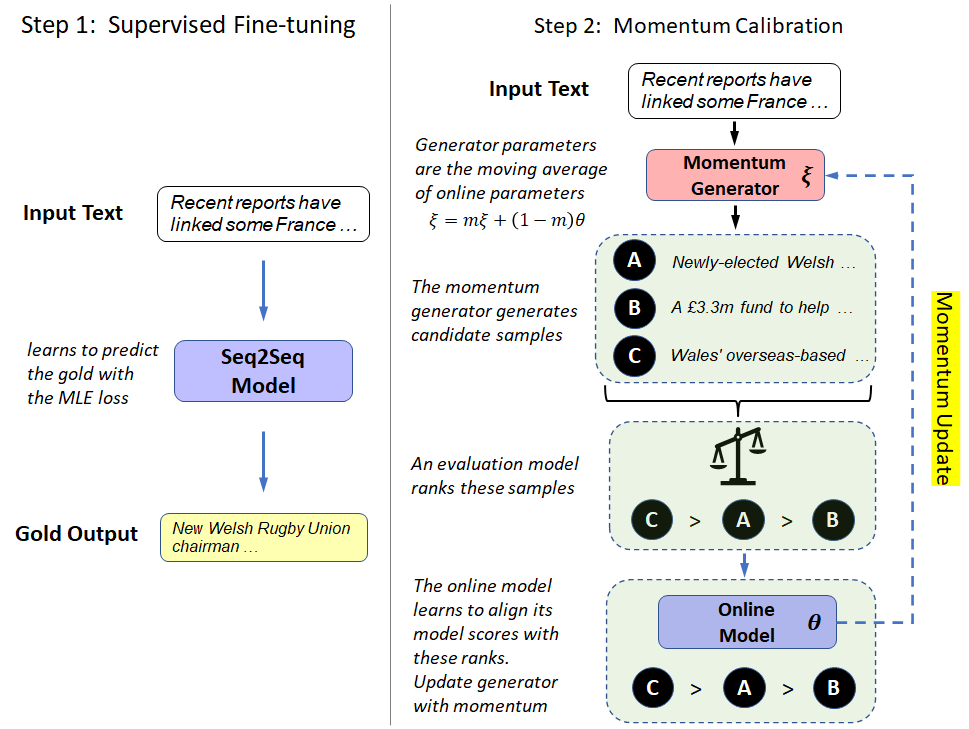}
    \caption{The training process of MoCa. Both the \emph{generator} and \emph{online} models are initialized from a supervised fine-tuned model. The momentum \emph{generator} first generates candidate samples. Then these samples are evaluated by an evaluation model and a ranked list is obtained. There may be a discrepancy between \emph{online} model scores and ranks of these samples. A ranking loss is used to adjust the \emph{online} model. Finally the momentum update is applied to the \emph{generator}.}
    \label{fig:moca}
\end{figure}

We introduce our model MoCa ({\bf Mo}mentum {\bf Ca}libration for Text Generation) in this section. We first introduce the vanilla training method for Seq2Seq models using maximum likelihood estimation, which suffers from \emph{exposure bias}. Then we introduce momentum calibration to cope with  this issue.

\subsection{Neural Text Generation}
\label{sec:generation}
Neural text generation aims to generate a text sequence according its input, which can almost always be converted to another text sequence. We usually adopt a sequence-to-sequence (Seq2Seq) Transformer \cite{Vaswani:2017:NIPS} to model this task. Given an input text sequence $X=( x_1, x_2, \ldots, x_{|X|} )$ and its gold output text sequence $Y= (y_1, y_2, \ldots, y_{|Y|} )$, the conditional probability $p(Y \mid X ; \theta)$ is estimated as follows:
\begin{equation}
    \label{eq:model_prob}
    p(Y \mid X ; \theta)=\prod_{t=1}^{|Y|} p\left(y_t \mid y_{<t}, X ; \theta\right)
\end{equation}
where $\theta$ is the model parameter and $y_{<t}$ stands for all tokens before time step $t$.
The model can be trained by minimizing the negative log likelihood of input and output text pairs (equivalent to maximizing the likelihood):
\begin{equation}
    \label{eq:nll_loss}
    \mathcal{L}^{\text{MLE}}(\theta)=-\frac{1}{|Y|} \log p(Y \mid X ; \theta)
\end{equation}

During test time, the model is expected to generate the whole sequence of text from scratch using beam search, which is different from its training (i.e., predicting the next token given previous \emph{gold} tokens). Thus, the model is never exposed to its own error during training and this is called 
\mbox{\emph{exposure bias}} \cite{ranzato:2015:arxiv}. We introduce how to address exposure bias in the following.

\subsection{Momentum Calibration}
Assuming that we already have a Seq2Seq Transformer model trained using the MLE objective described in Section \ref{sec:generation}. So that the model can produce reasonable outputs. To cope with the \emph{exposure bias} problem encountered using beam search during inference, as shown in Figure \ref{fig:moca} we first generate output samples from the pre-trained generator model, which represent the model distribution. Then, we evaluate these samples with an evaluation model and obtain ranks of these samples. Usually, these ranks w.r.t. the evaluation model is different from ranks w.r.t. the model probabilities and that is why we need to do the calibration. We therefore \emph{force} the model probabilities to be aligned with the evaluation model outputs using a ranking loss. Finally we update the generator model and our method is online. 

In MoCa, we have two Seq2Seq Transformer models: the \emph{online} model $M(\theta)$ with parameters $\theta$ and the \mbox{\emph{generator}} model $G(\xi)$ with parameters $\xi$. They share the same model architecture, but they have their own parameters. At the beginning of training, we set $\xi = \theta$.

\paragraph{Sample Generation} Our method aims to simulate the beam search inference during training. Therefore, given the input sequence $X$, we first generate $K$ samples $\tilde{Y}_1, \tilde{Y}_2, \dots, \tilde{Y}_K$ from our \mbox{\emph{generator}} $G(\xi)$ using beam search (BS) or its variant diverse beam search (DBS) \cite{vijayakumar:2016:arxiv} and these samples will be subsequently used to fix \emph{exposure bias}. In addition to the normalized log probabilities used in beam search, diverse beam search also take the differences between beam groups into account and it generates samples with a good trade-off between quality and diversity. We did not use sampling or nucleus sampling \cite{holtzman:2019:arxiv}, because the quality of samples generated by them are not as good as BS and DBS (the upper bound, mean and lower bound of ROUGE for samples are all lower). Besides, we also find that a significant portion of generate samples from nucleus sampling are duplicated. 

\paragraph{Evaluation and Calibration} We use samples generated above to calibrate our online model $M(\theta)$ in this step. Once we obtain these samples $\tilde{Y}_1, \tilde{Y}_2, \dots, \tilde{Y}_K$, we can evaluate these samples against the gold output sequence $Y$ with the evaluation model $\mathcal{E}(\tilde{Y}_k, Y)$. $\mathcal{E}$ can be a non-parametric model such as ROUGE \cite{lin-2004-rouge} or BLEU \cite{papineni-etal-2002-bleu} and/or a parametric model like BERTScore \cite{zhang2019bertscore}. Then we rank these samples w.r.t. the evaluation model and obtain a list of ranked samples $\tilde{Y}_1', \tilde{Y}_2', \dots, \tilde{Y}_K'$ so that $\mathcal{E}(\tilde{Y}_i', Y) < \mathcal{E}(\tilde{Y}_j', Y) \quad \forall \: i < j$.

Intuitively, a good model should assign higher probabilities to samples with higher evaluation scores. Our model $M(\theta)$, which is initially trained by the MLE loss (see Section \ref{sec:generation}), do not always assign samples probabilities aligned with evaluation scores. We use the following margin based pairwise ranking loss \cite{hopkins2011tuning,zhong-etal-2020-extractive} to adjust our model so that it can rank samples with high evaluation scores higher:
\begin{equation}
    \label{eq:rank}
    % \small
    \mathcal{L}^{\text{R}}(\theta) = \sum_{i < j} max( 0, s_{\theta}(X, \tilde{Y}_j') - s_{\theta}(X, \tilde{Y}_i') + (j-i) * \lambda )
\end{equation}
where $s_{\theta}(X, \tilde{Y}')$ is the  score assigned by model $M(\theta)$ for input $X$ and one of the samples $Y'$. $(j-i) * \lambda$ is the \emph{dynamic} margin between scores of $\tilde{Y}_i'$ and $\tilde{Y}_j'$ and $\lambda$ is a hyper-parameter.

Since our calibration process simulates the beam search search inference, naturally  $s_{\theta}(X, \tilde{Y}')$ is defined as the normalized log probabilities (i.e., the scoring function used in beam search):
\begin{equation}
    \label{eq:onlinescore}
    % \small
    s_{\theta}(X, \tilde{Y}') = - \frac{1}{ {|\tilde{Y}'|}^{\alpha} } \sum_{t=1}^{|\tilde{Y}'|} \gamma_t \log p\left(\tilde{y}_t' \mid \tilde{y}_{<t}', X ; \theta\right)
\end{equation}
where $\tilde{Y}' = (\tilde{y}_1', \tilde{y}_2', \dots, \tilde{y}_{|\tilde{Y}'|}')$, $\alpha$ is a hyper-parameter (similar to the length penalty in beam search) and the positional weighting function is constant (i.e., $\gamma_t=1$) in vanilla beam search.

Given a model trained by the MLE loss, we observe that 
the model usually tends to have lower word level predictive accuracy (w.r.t. gold) at later positions. Probably because the model needs to memorize more proceedings tokens to make the prediction.

In order to let the model focus on later positions (where the model tends to make mistakes), we propose a monotonically increasing weighting function for different positions.

\begin{equation}
    \tilde{\gamma_t} = \frac{1}{ ( |\tilde{Y}'| + 1 - t )^2 }
\end{equation}
We use the above function because $\sum_{t} \tilde{\gamma_t}$ has an upper bound and there is no additional hyper-parameter in it. Note that $\lim_{n\to \infty} \sum_{i=1}^{n} \frac{1}{i^2}=\pi^2/6$.

% We also define another schedule for $\gamma$
To make sure the resulting $s_{\theta}(X, \tilde{Y}')$ and the $s_{\theta}(X, \tilde{Y}')$ used in vanilla beam search (i.e., $\gamma_t=1$)
are in the same scale, we normalize $\tilde{\gamma_t}$ with their mean and the resulting positional weighting function is:

\begin{equation}
    \label{eq:scoring2}
    \gamma_t = \frac{ \tilde{\gamma_t} }{\frac{1}{|\tilde{Y}'| } \sum_{i=1}^{|\tilde{Y}'|} \tilde{\gamma_i} } =  \frac{|\tilde{Y}'|}{(|\tilde{Y}'| + 1 - t)^2 \sum_{i=1}^{ |\tilde{Y}'| } \frac{1}{i^2} }
\end{equation}
Note that the above positional weighting function is used when positional accuracies of the model trained with MLE loss drops in later positions. Otherwise, we use the constant positional weighting function $\gamma_t = 1$.

In the final model loss, we put a small weight on the MLE objective (see Equation \ref{eq:nll_loss}), and we intend to remind the model what has been learned in the first stage MLE training (Section \ref{sec:generation}):
\begin{equation}
    \label{eq:final_loss}
    \mathcal{L}(\theta) = \mathcal{L}^{\text{R}}(\theta) + \beta  \mathcal{L}^{\text{MLE}}(\theta)
\end{equation}

\paragraph{Momentum Update} Using the training loss in Equation (\ref{eq:final_loss}), parameters of the \emph{online} model $M(\theta)$ can be updated. But the parameters of the \mbox{\emph{generator}} model $G(\xi)$ remains constant, since the beam search process is not differentiable. However, keeping $\xi$ constant is not reasonable, since as the training progresses and $M(\theta)$ becomes stronger, predicting the correct rankings of candidate samples from $G(\xi)$ might be too easy for $M(\theta)$. Indeed, in experiments we observed the model converges very fast (usually within one epoch), when keeping $M(\theta)$ unchanged. 

An alternative is to reset $\xi = \theta$ after each model update. In this case, our method becomes a \emph{fully} online method. Perhaps due to the rapid changing of the \emph{generator} $G(\xi)$ and its resulting  samples, the training loss is high (maybe because it is too difficult to learn) and we did not obtain very good results. 

To overcome the fast convergence and training instability problems, we finally opt for the momentum update for parameters of the \emph{generator} $G(\xi)$:
\begin{equation}
    \xi \leftarrow m \xi + (1-m) \theta
\end{equation}
where $m$ is the momentum coefficient. Note that only $\theta$ is updated in back-propagation. We observe in experiments that a relatively large momentum coefficient is required (e.g., $m=0.99$), which indicates the stability of the \emph{generator} $G(\xi)$ is important.

\section{Experiments}

\subsection{Datasets}
We conducted our initial experiments on four different text generation datasets across different domains and with different input and output lengths. They are CNN/DailyMail (CNNDM; \citealt{nallapati2016abstractive}), XSum \citep{narayan2018don}, SAMSum Corpus \cite{gliwa-etal-2019-samsum} and Gigaword \citep{napoles2012annotated}.

\paragraph{CNNDM} contains online news articles paired with associated highlights (i.e., summaries) from the CNN and DailyMail websites. Following the pre-processing steps introduced in \cite{see-etal-2017-get},  287,227 document-summary pairs for training, 13,368 for validation and
11,490 for test are obtained.

\paragraph{XSum} aims to create a one-sentence professionally written summary for a given article \cite{narayan2018don}. The articles are collected from BBC online and their summaries are extremely abstractive. We use the official splits, which includes 204,045 examples for training, 11,332 for validation and 11,334 for test.

\paragraph{SAMSum} is a human-annotated dialogue summarization dataset, which is created by asking linguists to create messenger-like conversations and then asking another group of linguists to write summaries. It includes 14,732/818/819 dialogue-summary pairs for training/validation/test.

\paragraph{Gigaword} is a sentence compression dataset, which is created from the Gigaword annotated English news corpus \cite{napoles2012annotated} and use the first sentence and the abstractive headline from a given article as a input-output text pair. We follow the same pre-processing steps in \cite{rush2015neural} and  obtain 3,803,957 sentence-pairs for training, 189,651 for validation and for 1,951 test.

\subsection{Implementation}
We use PEGASUS \cite{zhang:2019:icml} (568M parameters) as our backbone on XSum and BART \cite{lewis-etal-2020-bart} (400M parameters) is the backbone on other datasets. We use Adam \cite{kingma2014adam} to optimize our model and the learning rates and warmup steps are tuned on the validation sets. We generate 16 candidate samples using diverse beam search \cite{vijayakumar:2016:arxiv} during training. The margin coefficient $\lambda$ in Equation (\ref{eq:rank}) is set to 0.001. We use ROUGE score \cite{lin-2004-rouge} as our evaluation model, since it is faster to compute than model based methods such as BERTScore \cite{zhang2019bertscore}. The length normalization term $\alpha$ in our online model score function is set to 0.6 on XSum and 2.0 on other datasets. We set the weight for the MLE loss $\beta$ to 0.01 (Equation \ref{eq:final_loss}). In general, we find a large momentum coefficient $m$ (e.g., $m \ge 0.99$) is needed and larger momentum should be used for datasets with slower convergence.
We set $m=0.995$ on CNNDM, and $m=0.99$ is used on other datasets.

\subsection{Results}

\begin{table}[t]
\small
\begin{center}
\resizebox{\textwidth}{!}{
\begin{tabular}{l|l|c|c|c|c}
\hline

\multirow{2}{*}{\bf Models} & \multirow{2}{*}{\bf |$\theta$|} & CNNDM & XSum & SAMSum & Gigaword \\
 & & ROUGE-1/2/L & ROUGE-1/2/L & ROUGE-1/2/L & ROUGE-1/2/L \\

\hline
\hline

PEGASUS \cite{zhang:2019:icml}
& 568M
& 44.17/21.47/41.11
& 47.21/24.56/39.25
& --
& 39.12/19.86/36.24
\\

BART \cite{lewis-etal-2020-bart}
& 400M
& 44.16/21.28/40.90 
& 45.14/22.27/37.25 
& 53.42/28.14/49.03$^\dag$
& 39.12/20.06/36.37$^\dag$
\\

Z-Code++ \cite{he2022z}
& 710M
& --  /22.2/ --
& --  /24.6/ --
& --  /30.3/ --
& --
\\

ST-MoE \cite{zoph2202st}
& 268B
& --   /21.7/ --
& {\bf --   /27.1/ --}
& --
& --
\\

\hline

BRIO \cite{liu-etal-2022-brio}
& 400M*
% & 44.16/21.28/40.90  
& 48.01/23.80/44.67
& 49.07/25.59/40.40
& 53.74/29.06/49.37$^\dag$
& 39.20/20.16/36.43$^\dag$
\\

SLiC \cite{zhao2022calibrating}
& 2B
&  47.97/24.18/44.88 
&  49.77/27.09/42.08
& 54.37/29.88/45.89
& -- 
\\

\hline
\hline

Finetuned$^\ddag$
& 400M*
% & 44.16/21.28/40.90
& 44.22/21.22/41.01
% & 47.21/24.56/39.25
& 47.14/24.53/39.31
& 53.42/28.14/49.03
& 39.12/20.06/36.37
\\

MoCa
& 400M*
& {\bf 48.88/24.94/45.76}
& 49.32/25.91/41.47
& {\bf 55.13/30.57/50.88}
& {\bf 39.63/20.57/36.78} \\

\hline

\end{tabular}
}
\end{center}
\caption{Results on CNNDM, XSum, SAMSum and Gigaword datasets. The |$\theta$| column describes the number of parameters for each model. * means for BRIO, Finetuned and MoCa, their parameters are 568M on XSum and 400M on other datasets. $\dag$ indicates results obtained using our implementation, since they are not reported in \cite{lewis-etal-2020-bart,liu-etal-2022-brio}. $\ddag$ indicates our own supervised fine-tuning results.}
\label{tab:main}
\end{table}

Our main results are shown in Table \ref{tab:main}. In the first block, we compare MoCa with large pre-trained transformers using vanilla fine-tuning. PEGASUS \cite{zhang:2019:icml} and BART \cite{lewis-etal-2020-bart} are pre-trained on unsupervised text data using gaped sentence prediction and text infilling objectives, which contains 400M and 568M parameters, respectively. Z-Code++ \cite{he2022z} (710M parameters) and ST-MoE \cite{zoph2202st} (268B parameters) all employ the corrupted span prediction objective \cite{raffel2020exploring} and Z-Code++ leverage an addition replaced token detection objective \cite{clark2020electra}. MoCa outperforms all of them except for ST-MoE on XSum, despite of having only 400M or 568M parameters. Note that ST-MoE are 470x larger than MoCa.
We also compare with models using \emph{advanced} fine-tuning in the second block. BRIO (400M or 568M parameters) \cite{liu-etal-2022-brio} and SLiC (2B parameters) \cite{zhao2022calibrating} also try to align the model scores with evaluation metrics as our method MoCa, but they are offline methods and their model scoring functions are different from ours. We use the same backbone models as BRIO and we consistently outperform it across all datasets, which indicates that frequently updating candidate samples and using beam search tailored scoring functions are important. SLiC are around four times larger than our model, we still outperform them on CNNDM and SAMSum. Compared to our own implemented vanilla fine-tuning method (Finetuned) in the third block, MoCa outperform it by a significant margin across all datasets, which demonstrates MoCa can be a good replacement for vanilla fine-tuning. 

\subsection{Ablation}

\begin{figure}
     \centering
     \begin{subfigure}[b]{0.47\textwidth}
         \centering
         \includegraphics[width=\textwidth]{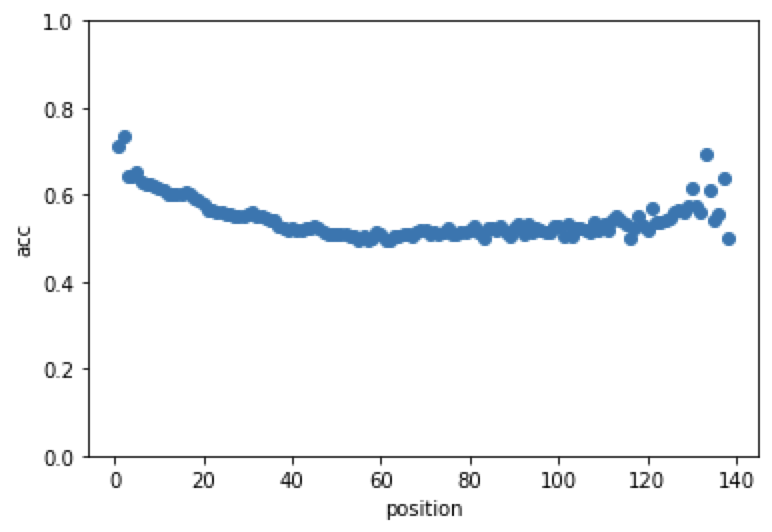}
         \caption{Positional accuracies on CNNDM.}
         \label{fig:pos_acc_cnndm}
     \end{subfigure}
     \hfill
     \begin{subfigure}[b]{0.47\textwidth}
         \centering
         \includegraphics[width=\textwidth]{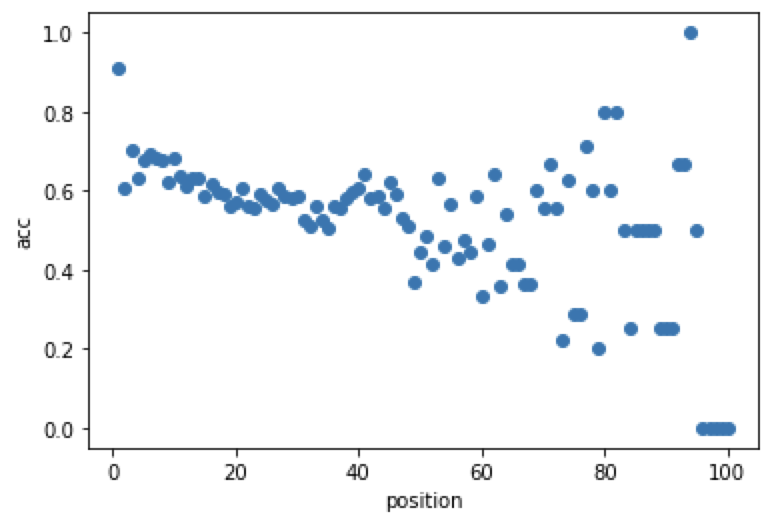}
         \caption{Positional accuracies on SAMSum.}
         \label{fig:pos_acc_samsum}
     \end{subfigure}
        \caption{Positional accuracies of models finetuned with the MLE loss on CNNDM and SAMSum datasets. The positional indices are measured using number of tokens by far in gold target sequences.}
        \label{fig:pos_acc}
\end{figure}

In this section, we assess the effectiveness of our proposed online model scoring functions and the momentum update strategy on the CNNDM and SAMSum datasets. When we use vanilla online scoring function ($\gamma_t = 1$) and do not use the online momentum update for the \emph{generator} model, our method is similar to BRIO \cite{liu-etal-2022-brio} (see the offline (BRIO) row in Table \ref{tab:ablation}). Note that our re-implementation of BRIO achieves better results than \cite{liu-etal-2022-brio} (also see Table \ref{tab:main}). As shown in Table \ref{tab:ablation}, with the positional weighting function we proposed (Equation  \ref{eq:scoring2} in Section \ref{sec:model}), we obtain much better results on SAMSum, while worse results on CNNDM (offline + scoring). As mentioned in Section \ref{sec:model}, the positional weighting function is designed to fix the positional accuracy drop problem in models trained by the MLE loss (where MoCa is initialized from). As shown in Figure \ref{fig:pos_acc}, the positional accuracies on CNNDM are stable across different positions, while the positional accuracies on SAMSum are dropping (especially from position 0 to 50) as positional indices become larger.
% {\bf show results here.}
Therefore, we use the weighting function in Equation (\ref{eq:scoring2}) on datasets (e.g., SAMSum), which suffer from the positional accuracy problem and the constant positional weighting function ($\gamma_t = 1$) on other datasets (e.g., CNNDM).

When upgrading the offline method to online, we find momentum update is important. Pure online method without momentum (online w/ $m=0$) sometimes hurts, while online method with momentum (momentum) consistently improves over its offline counterpart. We have observed that one scoring function may perform better than the other. With the \emph{proper} positional weighting function, the momentum method can be further improved (MoCa).

\begin{table}[t]
\small
\begin{center}
%\resizebox{0.48\textwidth}{!}{
\begin{tabular}{l|c|c}
\hline
% & \multirow{2}{*}{\bf # Para.}
\multirow{2}{*}{\bf Settings}  & CNNDM & SAMSum  \\
 & ROUGE-1/2/L & ROUGE-1/2/L \\

\hline
MoCa & 48.88/24.94/45.76 & 55.13/30.57/50.88 \\
\hline
offline (BRIO) & 48.31/24.38/45.12 & 53.74/29.06/49.37 \\
% offline + scoring 1 & 48.13/24.13/44.91 & -- \\
offline + scoring & 45.79/22.55/42.48 & 55.07/29.75/50.90 \\
online w/ $m=0$ & 48.30/24.51/45.15 & 53.70/29.00/49.33 \\
momentum & 48.88/24.94/45.76 & 54.10/29.17/50.67 \\

\hline

\end{tabular}
\end{center}
\caption{Effects of the proposed positional weighting function and momentum update on CNNDM and SAMSum.}
\label{tab:ablation}
\end{table}

\section{Conclusions}
We propose MoCa for text generation, which is an online method and aims to resolve the discrepancy between model probabilities assigned to candidate samples and their quality. Experiments across different datasets show MoCa consistently improves upon vanilla fine-tuning with the MLE loss for large pre-trained transformers. We also demonstrate the importance of being online and use scoring functions tailored for beam search. MoCa is currently applied to English text generation tasks. We hope our method can be useful for multi-lingual text generation tasks such as machine translation and cross-lingual text summarization as well as generation tasks beyond text (e.g., text-to-image generation and text-to-speech synthesis). 

\bibliography{custom}
\bibliographystyle{plainnat.bst}

\end{document}